\documentclass[10pt,conference,a4paper]{IEEEtran}

\usepackage{times}
\usepackage{epsfig}
\usepackage{graphicx}
\usepackage{amsmath}
\usepackage{amssymb}
\usepackage{url}
\usepackage[ruled,vlined]{algorithm2e}
\usepackage{algorithmic}
\RequirePackage{booktabs} 

\ifCLASSINFOpdf
\else
\fi

\begin{document}

\title{Iterative  Bounding Box Annotation for Object Detection}

\author{\IEEEauthorblockN{Bishwo Adhikari}
\IEEEauthorblockA{ 
Tampere University, Finland\\
bishwo.adhikari@tuni.fi}
\and
\IEEEauthorblockN{Heikki Huttunen}
\IEEEauthorblockA{
Tampere University, Finland\\
heikki.huttunen@tuni.fi} 
}


\maketitle

\begin{abstract}
Manual annotation of bounding boxes for object detection in digital images is tedious, and time and resource consuming. In this paper, we propose a semi-automatic method for efficient bounding box annotation. The method trains the object detector iteratively on small batches of labeled images and learns to propose bounding boxes for the next batch, after which the human annotator only needs to correct possible errors. We propose an experimental setup for simulating the human actions and use it for comparing different iteration strategies, such as the order in which the data is presented to the annotator. We experiment on our method with three datasets and show that it can reduce the human annotation effort significantly, saving up to 75\% of total manual annotation work. 
\end{abstract}

\IEEEpeerreviewmaketitle

\section{Introduction}

Object detection is one of the core research fields in machine learning and computer vision. Recently, object detection algorithms have matured enough to solve real-world vision problems. It serves as the key component in application fields such as face detection \cite{Hjelms2001FaceDA}, pedestrian detection \cite{pedestrian}, surveillance systems \cite{Surveillance_applications}, autonomous vehicles \cite{kitti_dataset, distant_vehicle_detection}, etc. The supervised learning principle is widely used in current object detection systems, where a human labeled dataset is used for training the detection model. The performance of supervised machine learning models relies heavily on the amount and quality of annotated training data. However, the challenge in supervised object detection is collecting large, high-quality labeled datasets with the aim of having a well-performing object detection model.  

Recently, both the scale and the variety of public datasets for object detection has increased. Among the most popular ones are PASCAL VOC \cite{pascal_voc}, MS COCO \cite{coco_dataset}, OpenImages \cite{OpenImages}, and Kitti \cite{kitti_dataset} and they are widely used as benchmark datasets in the field of object detection. The labor-intensive and tedious job of object annotation for these large datasets has often been solved by crowdsourcing \cite{Crowdsourcing} a large number of human annotators on web platforms such as Amazon Mechanical Turk. However, crowdsourcing may not be a feasible option for annotating small and medium-sized datasets, when the data is confidential in nature, or simply when annotation resources are limited. Hence, there is demand for resource-efficient, user-friendly annotation tools to prepare labeled dataset for machine learning.

Researchers have been mainly focusing on two approaches to reduce the cost of bounding box annotation, weakly-supervised and active learning methods. The weakly-supervised approach uses images and corresponding object labels and lets the network draw bounding boxes. On the other hand the active learning approach trains the model and requests human to draw bounding boxes on a subset of images actively selected by the learner itself. These approaches still require a significant amount of human annotator time for drawing high-quality bounding boxes.

In this paper, we present a simple and practical heuristic to annotate the bounding boxes on image datasets. The iterative annotation approach takes advantage of the trained model to propose labels for a batch of unlabeled images leaving the annotator only for correction work. Thus reducing the workload of annotation. Compared to other approaches, the iterative approach alternating between training-prediction-annotation balances in the work between machine and annotator for training and correction. 

Although commercial and open source tools that learn while annotating do exist (\textit{e.g.,} hasty.ai \cite{hasty} and ilastic \cite{berg2019}), there is only a limited amount of academic research on the topic. Moreover, our particular focus is not on tool development, but rather a systematic study investigating different \textit{strategies} for an annotation campaign; in particular the order in which the images are presented to the annotator. To this aim, we describe an easy-to-use measure of workload based on precision and recall metrics of the learnt machine learning model.

\begin{figure*}[t]
    \centerline{\includegraphics[scale=0.72]{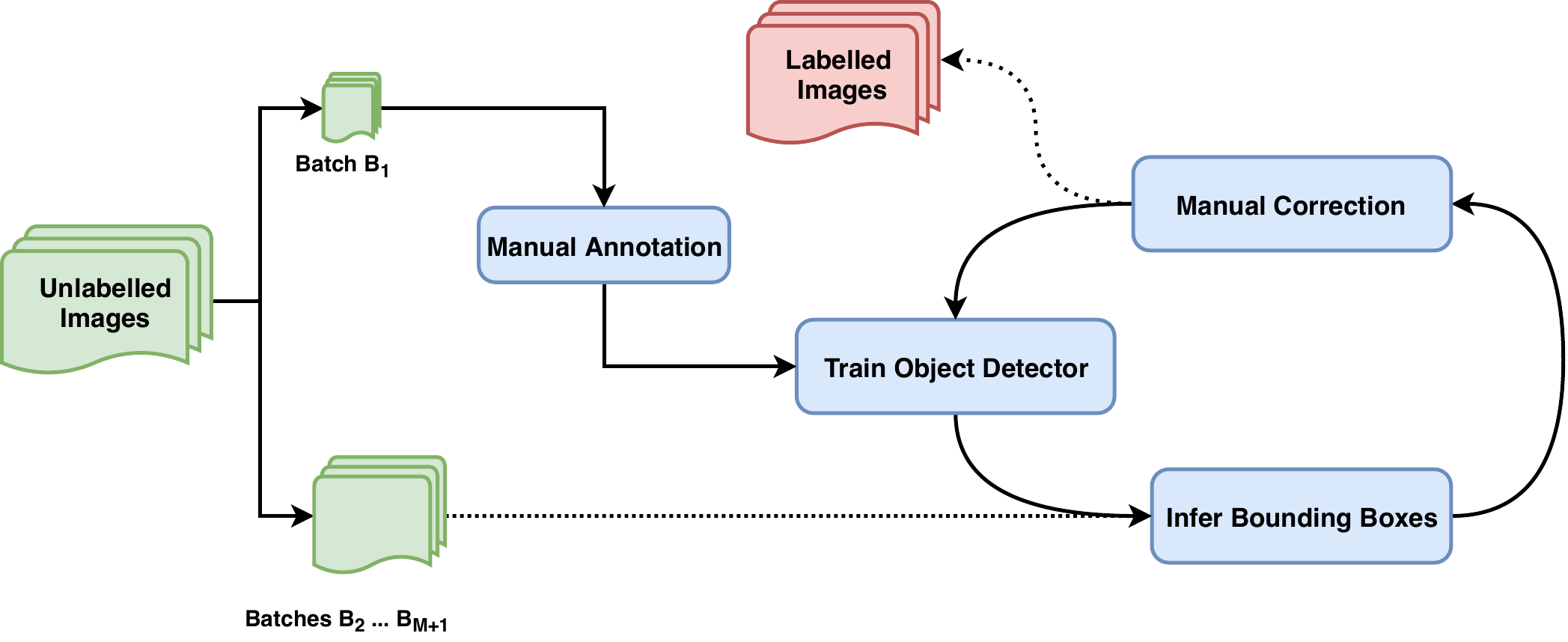}}
    \caption{Our purposed system for iterative bounding box annotation. The object detection model is trained on a small batch of manually annotated images. The trained model is used to predict labels on an unlabeled batch, followed by manual correction. After the first round of \textit{train-infer-correction}, detector is trained on the recently labeled batch. This process continues in a loop until all unlabeled batches are labeled. }
\label{method_pic}
\end{figure*}

Our proposed strategy significantly reduces the workload to create environment specific object detection datasets. Iterative training strategy is handy for a data annotation campaign, reduces tedious manual work by assisting annotators in real-time. A single annotator can efficiently annotate whole dataset utilizing a partly trained detector with our iterative training -- labels proposal method. We experiment with the \textit{continual learning} \cite{catastrophic_forgetting} effect, an ability of network to learn consecutive tasks without forgetting how to perform on previously trained task, on object detection models in an iterative loop. Additionally, the \textit{catastrophic-forgetting} \cite{catastrophic_forgetting} behaviour of object detection models are experimented with multiple approaches. The term \textit{catastrophic forgetting} resembles neural networks tendency of forgetting knowledge from the former data after learning from the new data.

The rest of this paper is structured as follows. We review the related literature on object annotation in Section~\ref{sec:related_work}. This is followed by a detailed discussion of the individual components of the proposed iterative annotation method in Section~\ref{sec:method}. Next, we present our experimental setup with a brief description of the dataset and pre-processing approaches together with the discussion of experimental results in Section~\ref{sec:experiment}. Finally, in Section~\ref{sec:conclusion}, we conclude our work and consider potential research directions for the future.

\section{Related Work} 
\label{sec:related_work} 

Object annotation in digital images has been widely studied since the first object detection methods were proposed in computer vision. 
There have been many studies focusing on speeding up the image dataset annotation for object detection task. In \cite{box_verification, click_supervision, Extreme_clicking}, Papadopoulos \textit{et al.} proposed multiple approaches for bounding box annotation. In their bounding box verification approach \cite{box_verification}, annotator only needs to verify the label proposed by the network with an accept/reject decision by  human. In the click supervision approach \cite{click_supervision}, the human annotator marks the point in the center of object in an image; and in the extreme clicking approach \cite{Extreme_clicking}, annotator clicks on four physical points on the object: the top, bottom, left- and right-most points. 

Among these works, the box verification approach \cite{box_verification} is the most similar to the proposed one. However, the proposed approach is different from all of these methods in that (1) we treat the object detector as a whole instead of splitting the task into object proposal and classification stages; (2) our human involvement is different (\textit{i.e.,} bounding box \textit{correction} instead of \textit{verification}); (3) we evaluate the performance in terms of total workload; and (4) we also study the order in which the examples are presented to the annotator directing the research towards active learning.

Konyushkova \textit{et al.} proposed learning intelligent dialogs \cite{learning_intelligent_dialogs}  that takes  advantage of a trained network to a draw bounding box on image. It requires human annotator to verify the bounding box proposed by the detector in all images. Again, our human interaction model is more straightforward (correction instead if yes/no verification), which in fact simplifies the task and the total workload (as most proposals need no action). 

In our previous paper \cite{adhikari_2018}, we used a two-stage semi-automatic approach to speed up bounding box annotation on labeling small training dataset and correction of network proposals. The proposed approach is related, but we extend the two-stage approach into an iterative training loop with unlimited number of training iterations rather than just two. 

There are commercial annotations tools available, such as hasty.ai \cite{hasty} and ilastic \cite{berg2019}. Although these tools exist already, the semi-automatic annotation approach \textit{specifically in bounding box annotation} has not been studied in the literature. To the best of our knowledge, this is the first systematic study of how different strategies could work in manual annotations workload reduction.

\section{Method}
\label{sec:method}
Our focus is in the iterative annotation framework illustrated in Figure~\ref{method_pic}. This annotation framework uses an incremental learning approach on a small batch of manually labeled images, trains a detection model, uses freshly trained model to propose bounding boxes on a batch of unlabeled images, and requests the annotator do the correction on possible incorrect bounding boxes or labels proposals.
The involvement of human annotators is only in the correction stage, hence, decrease the tedious task of manual annotation. Algorithm \ref{alg:algo} summaries all steps of the iterative training method. Next, we will describe each component of our method.

\begin{algorithm}
    \caption{Iterative annotation }
    \label{alg:algo}
    \begin{algorithmic}[1]
        \REQUIRE Set of unlabeled images split to $M +1$ distinct annotation batches $B_0, \ldots, B_{M+1}$
        \STATE annotate images in batch $B_0$ manually
        \STATE train object detection model with images from $B_0$
    \FOR {$i \in 1,2,\ldots,M$} 
        \STATE propose annotations for batch $B_i$ using the current prediction model 
        \STATE do manual correction for the proposals
       \STATE fine-tune the object detection model with  batch $B_i$ 
    \ENDFOR
\end{algorithmic}
\textbf{return} fully labeled dataset
\end{algorithm}
 
\subsection{Manual Annotation} 
The first step is to fully annotate the first batch of (say, 50) images from the unlabeled dataset. This stage is fully manual and requires human involvement to draw bounding boxes and provide class label on images. We use a basic bounding box annotation tools with no extra speed up procedures. The ways of selecting images batch for manual annotation are described later in Section \ref{approaches}.

\subsection{Object Detection Model Training} 
The second step is to train the object detection model. Although any detector can be used, we focus on the recent deep learning-based object detection models. The common practice is to use a pre-trained network and fine-tune on a new dataset  \cite{transfer_learning}. 
We choose two pre-trained networks trained on the MS COCO \cite{coco_dataset} dataset and fine-tune on other widely used datasets. Details of our training strategies are explained in Section \ref{sec:experiment}.

\subsection{Bounding Box Proposals} 
After fine-tuning the object detection model with the batches annotated so far, it is used to predict bounding boxes for the next batch of unlabeled images. The most recently trained detection model proposes bounding boxes and class labels on the next unlabeled batch of images.

\subsection{Manual Correction}
The proposed annotations are inspected and manually corrected by a human annotator. The human annotator needs to go through all the proposed labels and bounding boxes. Incorrectly predicted boxes are removed, wrongly labeled classes are corrected and new boxes are drawn, if needed. 
As the model is presented more samples during the iterations, the human workload should decrease, and the user only needs to accept the boxes in most cases. 

\subsection{Estimating the Workload}

Our goal is to estimate the human workload in a simulated setting, attempting to find out \textit{how much time a human would spend} on a full annotation campaign. Namely, we can compute the number of corrections required from the user with the commonly used precision and recall metrics. More specifically, the datasets are fully annotated, but we will process them iteratively to measure how much work a human would have done at each stage; and to compare different strategies how this could have been done faster.

The amount of overlap between the ground truth label and the predicted bounding box is used to define the false positives and false negatives. We assume the user would correct the annotation if the \textit{intersection-over-union} (IoU) overlap between the true object location and the predicted bounding box is less than 50\%, which is commonly used in performance evaluation in object detection. 

The formula to calculate the amount of corrections to bounding boxes and class labels is adopted from Adhikari \textit{et al.} \cite{adhikari_2018}. The number of \textit{additions} the user would have to perform for the next batch $B$ of images is given as
\[\text{\# additions} = (\text{\# of true objects})\times (1 - \text{recall}), \]
with the  recall metric computed from ground truth annotations for $B$. 
Moreover, the number of removals the user would have to perform (for false positives) is given as
\[ \text{\# removals} = (\text{\# of all detections})\times (1 - \text{precision}). \]
with the precision  computed from ground truth annotations for $B$. 
Finally, the total correction work is the sum of these two steps:
\[ \text{\# corrections} = \text{\# additions} + \text{\# removals}. \]

In an ideal case, when the detection model is good enough, the correction work on proposed bounding boxes should take significantly less time than drawing new boxes \cite{adhikari_2018}.
Additionally, in the case of partial overlap less than 50\%, we model the user operation as removal of the incorrect box and addition of the box at the correct location.

\subsection{ Labeled Dataset }
After every correction stage, the fully labeled batch is gathered and use for training the detection model for the next iteration. The cycle of training detection model, bounding box inference and correcting on proposed annotations loop continues until all unlabeled images are fully labeled. Hence, the loop produces a fully labeled image dataset for object detection in iterative loop with reduced workload for human annotator.

\section{Experiments and Results} 
\label{sec:experiment} 

In this section, we evaluate the efficiency of the proposed approach.
More specifically, we experiment on the iterative annotation with a number of detection architectures and datasets; both described below.

\subsection{Models}
\label{sec:models}

For detection architectures, we study the following two commonly used configurations. Both networks were trained starting with weights pre-trained with MS COCO dataset \cite{coco_dataset}.

\textbf{\em Faster RCNN---}The faster region convolutional neural network (Faster RCNN) framework proposed by Ren \textit{et al.} \cite{ren2015faster} is a two-stage detector, where the first stage network creates object proposals, followed by the second stage network classifying the proposals into different categories. For the backbone, we use the 50-layer variant of the residual convolutional network (Resnet50) \cite{he2016deep}. 

\textbf{\em SSD---}A commonly used lightweight detection architecture, the single shot detector (SSD) framework was proposed by Liu \textit{et al.}  \cite{liu2016ssd}. Since the detections are produced directly in a single forward pass of the network, it is often the model of choice for resource limited inference scenarios. For the backbone, we use the Mobilenet V2 \cite{sandler2018mobilenetv2}.

These particular models were selected since Faster RCNN is typically more accurate in detection, while the SSD is faster in inference. As detection model complexity plays crucial role in training, it is always worth to put some effort on model selection for the particular use case. Moreover, the idea is to experiment various strategies to find out an optimal strategy for high quality bounding box annotation efficiently in iterative approach.

\begin{figure*}[tb]
\centerline{\includegraphics[scale=0.4]{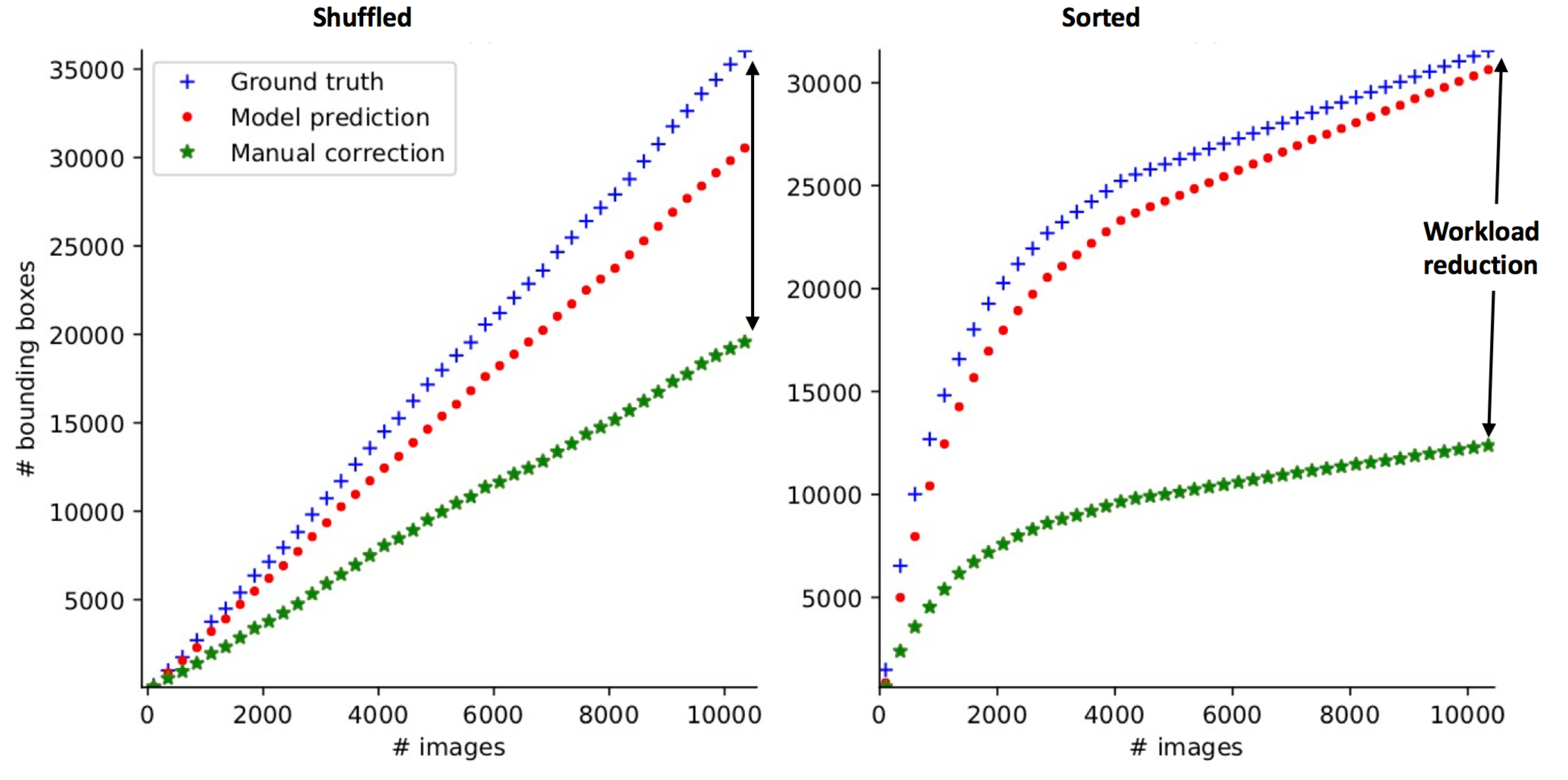}}
\caption{An example of the effect of the order of iterative annotation. The figures show the cumulative number of ground truth boxes, boxes predicted by the RCNN model, and the manual corrections required on the OpenImages/Person dataset. Images are annotated in a random order (left) and in an order sorted by the \# of boxes per image (right). The reduction in workload is significantly better in the sorted order (right).}
\label{result_image}
\end{figure*}

\subsection{Datasets}
\label{sec:dataset}

We selected three datasets for our experiments. The intention is to include both large scale datasets (OpenImages) as well as small scale sets (Indoor). The large datasets represent the upper bound in the size of an annotation campaign, while the small scale is the common setup occurring in practice.

\textbf{\em PASCAL VOC---}%
PASCAL VOC dataset \cite{pascal_voc} is a popular benchmark dataset for object detection. The dataset consists of 17k images having 40k object instances from 20 class categories, including person, bus, car, and motorbike. Two sets of independent experiments were conducted in this dataset: first with all classes and second with individual class categories. In the latter case, we used the top ten object categories from VOC 2012 dataset, listed in Table \ref{results_2}.

\textbf{\em OpenImages---}%
OpenImages \cite{OpenImages} is yet another very large dataset for object detection, classification and instance segmentation. The OpenImages V4 contains 9.2M images, 15.4M bounding boxes for 600 object classes. In our experiment, we used a subset of the Open Image Dataset: 10.5k images are selected from the person class; filtering the occluded, truncated, and groups of objects with a single label (multiple objects inside a single bounding box), as these annotations tend to be very noisy. 

\textbf{\em Indoor---}%
Indoor dataset \cite{adhikari_2018} is a  moderate size dataset collected from university indoor premises. The fully annotated object detection dataset consists of about 2,200 images and about 5,000 object instances. Images were extracted from a series of videos. Therefore, this dataset is also associated with a temporal order of samples, which we will exploit later.

 Table~\ref{table:datasets} summarizes the characteristics of these three datasets. Note that although all datasets used in our experiments are fully labeled, we investigate how their relabeling could benefit from training in the annotation loop. The scope of this research is to experiment how bounding box annotation cloud be done faster with minimal human involvement in an iterative-train loop.

\begin{table}[!h]
\caption{Comparison of Indoor \cite{adhikari_2018}, PASCAL VOC 2012 \cite{pascal_voc} and OpenImages \cite{OpenImages} datasets. Source, size, annotation features and usages are presented in rows respectively.}
\centering
\begin{tabular}{l|l|l}
\toprule
\textbf{Indoor} & \textbf{PASCAL VOC} & \textbf{OpenImages} \\ 
\midrule
Indoor videos & Online (Flickr)  &  Online (Flickr)\\  \midrule
 \begin{tabular}[c]{@{}l@{}}2.2k images \\  4.5k instances \\  7 classes\end{tabular}  
& \begin{tabular}[c]{@{}l@{}}17.1k images\\  40k instances  \\  20 classes  \end{tabular}    
& \begin{tabular}[c]{@{}l@{}}9.2 M images\\  15.4M instances \\  600 classes  \end{tabular}    \\  \midrule

 \begin{tabular}[c]{@{}l@{}} Fully labelled by\\ one annotator; \\ annotations are \\ of high quality \end{tabular}  
& \begin{tabular}[c]{@{}l@{}} Fully labelled by\\ multiple annotators;\\ annotations are of \\good quality \end{tabular}    
& \begin{tabular}[c]{@{}l@{}} Partial labelled \\ multiple annotators \&  \\machine generated labels; \\annotations contain noise\end{tabular}    \\  \midrule
 object detection                  
         &  \begin{tabular}[c]{@{}l@{}}classification, \\ detection, \\ segmentation \end{tabular}
         &  \begin{tabular}[c]{@{}l@{}}classification, \\ detection, \\ visual relationship \end{tabular}  \\
\bottomrule
\end{tabular}
\label{table:datasets}
\end{table}

\begin{table}[tb]
\caption{Reduction of manual workload (\%) with different strategies on Indoor, PASCAL  VOC 2012 and OpenImages datasets.
Numbers in bold represent the best performing approach on each section.
}
\centering
\begin{tabular}{l|c|c|c}
\toprule
\textbf{Network - Approach} & \textbf{Indoor } 
 & \begin{tabular}[c]{@{}c@{}}\textbf{PASCAL VOC }\end{tabular}
& \begin{tabular}[c]{@{}c@{}}\textbf{OpenImages}\\\textbf{Person}\end{tabular}\\
\midrule
RCNN - Shuffled  & \textbf{75.86}  & 18.40               & 45.62  \\  
RCNN - Sorted    & 56.97           & 20.93               &\textbf{60.05}\\
RCNN - Original  & 35.78           & \textbf{25.23}      & 45.73     \\ 
\midrule
SSD - Shuffled  &\textbf{47.38}   & 3.46                 & 20.28  \\ 
SSD - Sorted    & 31.58           & 5.66                 &\textbf{35.13} \\
SSD - Original  & 19.24           & \textbf{7.97}        & 20.04 \\ 
\bottomrule
\end{tabular}
\label{results_1}
\end{table}

\begin{table*}
\caption{Comparison of the manual workload reduction (\%) on individual class categories from PASCAL VOC 2012 dataset.}
\centering
\begin{tabular}{l||c|c|c|c|c|c|c|c|c|c||c}
\toprule
  &  \textbf{Airplane} & \textbf{Bird} & \textbf{Boat} & \textbf{Bottle} & \textbf{Car}& \textbf{Cat} & \textbf{Chair} &\textbf{Dog} & \textbf{Person} & \textbf{Plant} & \textbf{Average}\\
\midrule
RCNN - Shuffled     & 56.14    & 50.30  &\textbf{35.70}     & 44.49     & 51.96    & 55.34      & 29.31     & 57.87     & 44.61    & \textbf{38.72} & 46.44 \\  
RCNN - Sorted      & \textbf{62.07}    &\textbf{60.43}  & 35.65     & \textbf{46.68}    & \textbf{56.27}    & 59.53      & \textbf{32.44}     & 63.28    & \textbf{61.24}    & 32.75 & \textbf{51.03} \\
RCNN - Original   & 53.87    & 50.41  & 32.50     & 41.54     & 55.14    & \textbf{61.58}      & 29.30     &\textbf{61.38}    & 57.16    & 34.64 & 47.75 \\ 
\bottomrule
\end{tabular}
\label{results_2}
\end{table*}

\subsection{Order of annotation} 
\label{approaches}

For each iteration, we use a batch size of 50 images in all our experiments. The input image size of 1024x600 pixels is used to train all RCNN models and the input size of 300x300 pixels is used to train all SSD models. The following three annotation orders are experimented on the above mentioned datasets.

\textbf{\em Shuffled---}%
All images on the dataset are shuffled in random order and divided into small batches of 50 images. All these small batches have randomly selected images from the whole dataset. The human annotator manually annotates the first batch, then fine-tune the detection model on that batch, and remaining batches are used for annotation proposal in an iterative loop.

\textbf{\em Sorted---}%
In this approach images from the dataset are sorted in the decreasing order of number of objects per image. Images having more objects (in both single and multiclass cases) are presented first and so on. In this setup, the first batch of 50 images contains comparatively more objects than the last batch. Hence, it takes more time to annotate the first batch than in any other setups.

\textbf{ \em Original---}%
In our next approach, images are presented in the original order defined either by the filename (PASCAL VOC and OpenImages) or temporal order in a video (Indoor) or some other inherent way of ordering the examples. This is likely to differ from the shuffled order only for temporal sequences such as video, but we also experiment with this order with the non-temporal datasets for the sake of completeness.

\subsection{Results}
\label{sec:results}

An example of the progression of our method experimented on OpenImages person class using the RCNN detection model is shown in Figure \ref{result_image}. The amount of ground truth, model prediction, and manual correction in terms of numbers of bounding boxes as a function of the number of images are presented with shuffled and sorted approaches. 

The left graph shows the experimental result based on the shuffled order of images from the dataset, and the right graph shows the experimental result based on the sorted order. The manual workload is less in the sorted order approach; only 13,414 manual corrections would be needed instead of 33,573 ground truth bounding boxes. In the case of the random shuffle approach, 19,938 manual corrections would be required instead of 36,659 ground truth bounding boxes.

The higher the gap between ground truth and manual correction, the better the annotation performance in terms of manual workload reduction. The first stage manually annotated bounding boxes from the first batch ($B_0$) of images are not included in these graphs. For the sorted approach, the annotation workload for the very first batch is usually more than other approaches.

The reductions of manual workload required to correct the proposed bounding boxes in experimented approaches are shown in Table \ref{results_1}. The prediction performances of the two-stage RCNN model are comparatively better than those of the single-stage SSD model. The RCNN model capacity and higher input resolution for training images result in good proposals for bounding boxes and class labels. However, the two-stage models are computationally expensive compared to single-stage detection models like SSD. Moreover, the results follow the same pattern for both methods in all of our experimented approaches. The minimum amount of manual workload reduction noted is 3.46 \% with the SSD model on the VOC multiclass dataset with the shuffling approach. On the other hand, the maximum amount of reduction recorded is 75.86\% with the RCNN model on the Indoor dataset with the shuffling approach.

Interestingly, it is seen that shuffling images helps to improve the overall performance of the detection model only in the case where images are continuous frames of video (Indoor). As shown in Table \ref{results_1}, with the RCNN model, 75.86\% of manual work required to draw bounding boxes can be reduced by randomly shuffling the images from the indoor dataset. However, in the shuffling approach, there is a high probability of having every next image from the sequence in different batches. 
The catastrophic forgetting effect seems to be least in this approach hence, providing accurate proposals for bounding boxes and class labels. Moreover, the results on OpenImages and VOC datasets show that the shuffled approach is the worst performing among the compared approaches. 

For multi-class datasets, the annotations can be done one class at a time (iterating over all images for each class) or all classes simultaneously. Therefore, we experimented with the ten most populated classes of the VOC dataset with the RCNN detection model, with a single category fully annotated iteratively at a time. 
The results are shown in Table \ref{results_2}.  Interestingly, the workload reduction is doubled when annotations are done one class at a time (average reduction up to 51.03\%) compared to the multi-class iterative annotation (reduction 25.23\%). In terms of individual categories, the reduction ranges from 32.4\% (chair) to 62.1\%  (airplane). The likely reason for the improved performance is that the detection model does not have to learn away from the other categories and can focus on learning a single class at a time. Obviously, this approach requires that images are initially sorted by category: Otherwise we would unnecessarily present, \textit{e.g.,} non-airplane images while annotating the airplane class. 

Additionally, it is found that the sorted approach is optimal for the manual workload reduction in most of the single class categories. On the OpenImages person class, as shown in Table \ref{results_1}, the workload reduction with the sorted approach is higher than other setups. The reduction in manual workload with RCNN model is  60.05\% and with SSD model is 35.13\%. Most of the larger reductions in Table \ref{results_2} come from the sorted setup. 

On the other hand, by splitting images into different batches based on the filename or temporal order gives the best result for the multiclass dataset. The original order approach gives best performance for VOC multiclass and second best performance for VOC and OpenImages single class datasets.

\begin{table}[!ht]
\caption{Comparison of the manual workload reduction (\%) with Iterative annotation and Two-stage method on Indoor dataset \cite{adhikari_2018}. All experiments are done with faster RCNN ResNet101 and 0.5 IoU threshold is used.}
\centering
\begin{tabular}{l|c}
\toprule
\textbf{Approach} & \textbf{Reduction (\%)} \\
\midrule
Two-stage (5\%) \cite{adhikari_2018}      &  79.47   \\
Two-stage (6\%) \cite{adhikari_2018}      &  81.21   \\
Two-stage (8\%) \cite{adhikari_2018}      &  78.68   \\
Two-stage (10\%) \cite{adhikari_2018}     &  79.03   \\
Two-stage (20\%) \cite{adhikari_2018}     &  72.46   \\
Ours (iterative)   & 79.56  \\
Ours (cumulative)  & 80.56  \\
\bottomrule
\end{tabular}
\label{comparision}
\end{table} 

We performed some more experiments on the Indoor dataset to have a comparison with the previous method \cite{adhikari_2018}. 
In \cite{adhikari_2018}, a two-stage approach was used, with first fold annotated fully manually, and the second part using the proposals of a model trained with the first fold. The proportion of samples in the first and second folds is a tuning parameter.

Table \ref{comparision} shows the results of the two-stage approach \cite{adhikari_2018} for selected split ratios; with 6\% for the first fold and 94\% for the second fold reducing the workload the most (81.21\%).
Using the fully iterative approach of this paper results in a comparable accuracy (79.56\%), but does not require the selection of the split ratio. 

However, the sweet spot of 6\% is the \textit{only} split ratio exceeding the workload reduction of the proposed iterative approach. In fact, it is impossible to know the optimal split ratio \textit{a priori} for an unlabeled dataset; for example, the 5\% -- 95\% split is already almost 2\% worse. Thus, our iterative method is economical compared to the earlier two-stage approach. 

In addition to the iterative approach, we experimented on a modified version of the proposed method. Namely, instead of training on the most recent batch, one can train with all previously annotated batches at each iteration. This \textit{cumulative} approach was tested on the Indoor dataset and the results are shown on the last line of Table \ref{comparision}. 
It turns out that the cumulative approach is better compared to the iterative approach; however, its training becomes significantly slower due to larger amount of training data.

In general, the two-stage approach with other amounts of manually labeled images has less workload reduction than our iterative approach trained with the same approach and same model. 
The significant benefit iterative approach over this is that the model is improving over time; object proposals are getting better, hence require less correction work. Also, labeling a small batch of images is more relaxed and less error-prone compared to mass annotation.

\section{Conclusion}
\label{sec:conclusion}

This paper presented an iterative human-in-the-loop annotation method to minimize human involvement in image annotation campaigns for the task of object detection. The proposed method utilizes human-machine collaboration resulting in high-quality bounding box annotations on small and medium-sized datasets. Extensive experiments on the proposed method on three datasets show that our iterative annotation is able to reduce the manual workload by up to 75\%. However, the reduction of manual annotation workload appears to be dependent on the dataset size, image source, and object class categories. We also noted an adequate performance of proposed approaches on the already existing platform for the bounding box annotation campaign.

Our implementation uses a commonly used threshold value 0.5 IoU threshold threshold for detection performance measurement and accepting the label proposals. In our future work, we will consider the effect of requiring a higher IoU for the annotation results ($> 0.5$ IoU). In fact, we already experimented with a 0.8 IoU threshold, and discovered that the manual workload reduction is even higher since less manual correction work is needed on the good quality proposals. It would also be interesting to experiment on the effect of an IoU threshold on the network proposals and its tradeoff between correction workload. 
The resource constraints and latency of the detection model could be studied in the future with parameters such as batch size, training step, and input size of images for the model training. 

Furthermore, it would be nice to experiment on an active learning pipeline that could reduce human workload even more by  automatically selecting informative images for labeling on earlier iterations.  

\bibliographystyle{IEEEtran}
\bibliography{references}

\end{document}